\documentclass[11pt]{article}
\usepackage{graphicx} 
\usepackage{cite}
\usepackage[T1]{fontenc}  
\usepackage{newtxtext,newtxmath} 
 
\usepackage{amsmath,amssymb,amsfonts}
\usepackage[linesnumbered,ruled,vlined]{algorithm2e} 
\usepackage{tabularx} 
\usepackage[normalem]{ulem}
\usepackage{xcolor}
\usepackage[margin=1.2in]{geometry} 
\usepackage{booktabs}
\usepackage{subcaption} 

\usepackage{bm,mathtools, scrextend, array, setspace, textcomp, comment} 

\newcolumntype{M}[1]{>{\centering\arraybackslash}m{#1}} 
\def\BibTeX{{\rm B\kern-.05emF{\sc i\kern-.025em b}\kern-.08em
    T\kern-.1667em\lower.7ex\hbox{E}\kern-.125emX}}

\graphicspath{ {./images/} } 

\providecommand{\keywords}[1]
{
  \small	
  \textbf{\textit{Keywords---}} #1
}

\title{Proximal Vision Transformer: Enhancing Feature Representation through Two-Stage Manifold Geometry}
\author{
    Haoyu Yun \and Hamid Krim 
}
\date{}

\begin{document}

\maketitle
\begin{abstract}

The Vision Transformer (ViT) architecture has become widely recognized in computer vision, leveraging its self-attention mechanism to achieve remarkable success across various tasks. Despite its strengths, ViT's optimization remains confined to modeling local relationships within individual images, limiting its ability to capture the global geometric relationships between data points. To address this limitation, this paper proposes a novel framework that integrates ViT with the proximal tools, enabling a unified geometric optimization approach to enhance feature representation and classification performance. In this framework, ViT constructs the tangent bundle of the manifold through its self-attention mechanism, where each attention head corresponds to a tangent space, offering geometric representations from diverse local perspectives. Proximal iterations are then introduced to define sections within the tangent bundle and project data from tangent spaces onto the base space, achieving global feature alignment and optimization. Experimental results confirm that the proposed method outperforms traditional ViT in terms of classification accuracy and data distribution.

\end{abstract}
\section{Introduction}
\label{Intro}

In recent years, Vision Transformers (ViT)\cite{dosovitskiy2021imageworth16x16words} have emerged as a highly regarded architecture due to their remarkable performance across various computer vision tasks. ViT employs the self-attention mechanism to model the intricate relationships within input data\cite{devlin2019bertpretrainingdeepbidirectional}. However, the primary focus lies in modeling relationships within individual images, with the attention mechanism exclusively capturing interactions between different patches of the same image. This inherent design introduces limitations, as it fails to fully leverage the global relationships between data points, which are critical for understanding the overall structure of the data manifold. Consequently, optimizing ViT is constrained to local scopes, making it challenging to fully explore the geometric properties of the underlying data.

This paper introduces a novel framework that integrates the Vision Transformer (ViT) with the proximity operator \cite{10096421,bertocchi2020deepunfoldingproximalinterior} to establish a unified geometric framework, thereby enhancing feature representation capabilities and improving classification performance. Within this framework, ViT employs the self-attention mechanism\cite{vaswani2023attentionneed} to construct the tangent bundle of the manifold, where each attention head corresponds to a tangent space on the manifold. These tangent spaces provide linear approximations of the manifold's local structures, offering a multi-perspective local geometric representation of the input data.

Building on this foundation, the proximity operator is incorporated to further optimize these representations. By defining sections within the tangent bundle and projecting data from the tangent spaces onto the base space, the proximity operator achieves global alignment of features. This optimization process not only enhances intra-class consistency but also significantly improves inter-class separability. Specifically, the proximity operator optimizes a self-expressive loss function, enabling the reconstruction of each data point to incorporate relationships with other points on the manifold, thereby capturing the global geometric characteristics of the data.

By combining the local feature modeling of ViT with the global geometric alignment provided by the proximity operator, the proposed framework performs dual-level optimization. It locally preserves fine-grained structure within each sample and globally enhances feature coherence across samples. Experimental results demonstrate that this framework excels particularly on high-resolution datasets, highlighting its advantages in handling complex features and high-resolution images.

In summary, the main contributions of this paper include proposing a novel framework that integrates ViT's tangent bundle construction capability with the proximity operator's section definition and base space projection ability, thereby achieving unified geometric optimization. Furthermore, the paper provides a theoretical perspective rooted in manifold geometry, significantly enhancing feature representation capabilities through the use of tangent spaces, tangent bundles, and sections. Lastly, the proposed framework is validated through comprehensive experiments, demonstrating its superiority in terms of classification performance and data distribution characteristics.

The remainder of this paper is organized as follows: Section~\ref{works} summarizes the research background and progress related to this work, including Vision Transformer, manifold geometry, and optimization methods; Section~\ref{methodology} details the proposed framework, including the geometric principles and optimization processes of ViT and the proximity operator; Section~\ref{experiments} presents experimental results and analyses, along with comparisons to the baseline ViT; finally, Section~\ref{conclusion} concludes the paper and discusses future research directions.

\section{Related work}
\label{works}

\subsection{From Local Modeling to Global Geometry}

The Transformer\cite{vaswani2023attentionneed} model was initially introduced in the field of natural language processing (NLP)\cite{qin2024largelanguagemodelsmeet}, where it achieved remarkable success by leveraging multi-head self-attention to capture global dependencies within sequential data. Its innovative architecture quickly became the foundation for tasks such as machine translation, text generation, and classification, and its application has since expanded to computer vision.

In computer vision, the Vision Transformer (ViT) extended the Transformer paradigm from one-dimensional sequential data to two-dimensional image data. By dividing images into fixed-size, non-overlapping patches and processing these patches as input sequences, ViT pioneered a non-convolutional approach to vision tasks\cite{dosovitskiy2021imageworth16x16words}. With access to large-scale datasets such as ImageNet-21k\cite{DBLP:journals/corr/abs-2104-10972}, ViT demonstrated outstanding performance by effectively modeling the relationships between image patches using self-attention mechanisms. However, ViT faces notable limitations: its performance heavily relies on pretraining on large datasets; it primarily focuses on local relationships within individual images, lacking the capability to explore global geometric relationships across samples; and it exhibits suboptimal performance on small datasets or scenarios with feature position variations.

To address these limitations, various extensions of ViT have been proposed. For instance, Swin Transformer\cite{liu2021swintransformerhierarchicalvision} introduces localized window-based self-attention to enhance efficiency while maintaining global feature modeling. DeiT\cite{touvron2021trainingdataefficientimagetransformers} incorporates knowledge distillation to reduce dependency on large datasets. CrossViT\cite{chen2021crossvitcrossattentionmultiscalevision} and Tokens-to-Token ViT\cite{yuan2021tokenstotokenvittrainingvision} further optimize multi-scale feature fusion and representation capabilities. However, these approaches primarily focus on intra-image relationship modeling, leaving the exploitation of global geometric structures across samples largely unexplored.

This paper proposes a novel framework that integrates ViT with the proximity operator to overcome these challenges. Our method not only leverages ViT’s self-attention mechanism to construct the tangent bundle of the data manifold but also applies proximal tools to achieve global feature alignment across samples. Compared to existing methods, this framework demonstrates significant advantages in improving classification performance, enhancing feature consistency, and capturing global geometric relationships, providing a new direction for the extension of Transformer-based models.

\subsection{Proximity Operator}

The proximity operator is a powerful tool in optimization, widely used in non-smooth optimization, sparse representation, and high-dimensional data analysis\cite{bertocchi2020deepunfoldingproximalinterior,tang2021deeptransformmetriclearning}. Its fundamental objective is to decompose complex objective functions and address the non-smooth components through a simple approximation operation, thereby allowing efficient optimization to be achieved with proper convergence guarantees. Over the years, its applications have significantly expanded, ranging from traditional convex optimization problems to the design of regularization in deep learning models  and geometric optimization tasks, demonstrating its flexibility and robustness. Formally, the proximity operator of a function $g$ defined on Hilbert space $\mathcal{H}$, possibly scaled by a nonnegative factor $\lambda$, is defined as

\[
(\forall x \in \mathcal{H})\quad 
\operatorname{prox}_{\lambda g}(x) = \arg \min_{z\in \mathcal{H}} \left( \frac{1}{2} \|x - z\|^2 + \lambda g(z) \right),
\]
where \(g\colon \mathcal{H}\to ]-\infty,+\infty)\) is typically a non-smooth regularization function, such as the \(\ell_1\)-norm or other sparsity-inducing constraints. The existence and uniqueness
of $\operatorname{prox}_{\lambda g}$ is guaranteed if
$g$ is convex, lower-semicontinuous, and proper.
By introducing the proximity operator, optimization problems can be split in different components, simplifying the handling of complex objective functions and improving efficiency.

In the field of machine learning, the proximity operator has found extensive use in sparse representation and regularized models\cite{tang2020deeptransformmetriclearning}. Methods such as Elastic Net\cite{b61d584f-9c9c-3cf8-b866-214ec2216250} and LASSO\cite{51791361-8fe2-38d5-959f-ae8d048b490d}, for example, are often solved using  proximal tools. Furthermore, in the domain of sparse subspace clustering, the proximity operator is employed to optimize self-expressive models, achieving both sparsity and low-rank constraints. Beyond these applications, the proximity operator has been adapted for solving optimization problems on manifolds. By defining sections and projections, it serves as a tool to optimize the geometric consistency of data points on the manifold. For instance, in unsupervised learning tasks, it has been used to optimize self-expressive relationships among data points, facilitating subspace clustering.

This study further extends the application of the proximity operator by integrating it into the optimization process of the Vision Transformer (ViT), thereby proposing a unified geometric optimization framework. In this context, the proximity operator is used not only to define sections within the tangent bundle but also to achieve global geometric alignment across data points through feature projection. This approach departs from traditional applications of the proximity operator, which mainly target non-smooth optimization problems with a focus on sparsity constraints or low-rank issues. Instead, this study employs the proximity operator to optimize the geometric consistency of tangent bundle representations, enhancing global feature alignment. Additionally, while traditional methods are often confined to relatively simple convex optimization scenarios, this study integrates the proximity operator into ViT’s self-attention mechanism, enabling the optimization of feature representations within deep models and significantly enhancing the global modeling capability of Transformers. Moreover, the proximity operator's role is expanded from intra-sample regularization to modeling global geometric relationships, optimizing both intra-class consistency and inter-class separability to improve classification performance from a geometric perspective.

By leveraging the theoretical advantages of the proximity operator, this study provides new insights into its application within deep learning models, effectively addressing the limitations of existing transformers in capturing and modeling global geometric relationships. This innovative integration of the proximity operator into the ViT framework highlights its potential for advancing both feature representation and task performance in complex machine learning problems.

\section{Methodology}
\label{methodology}

\begin{figure}[ht]
  \centering
  \includegraphics[scale = 0.45]{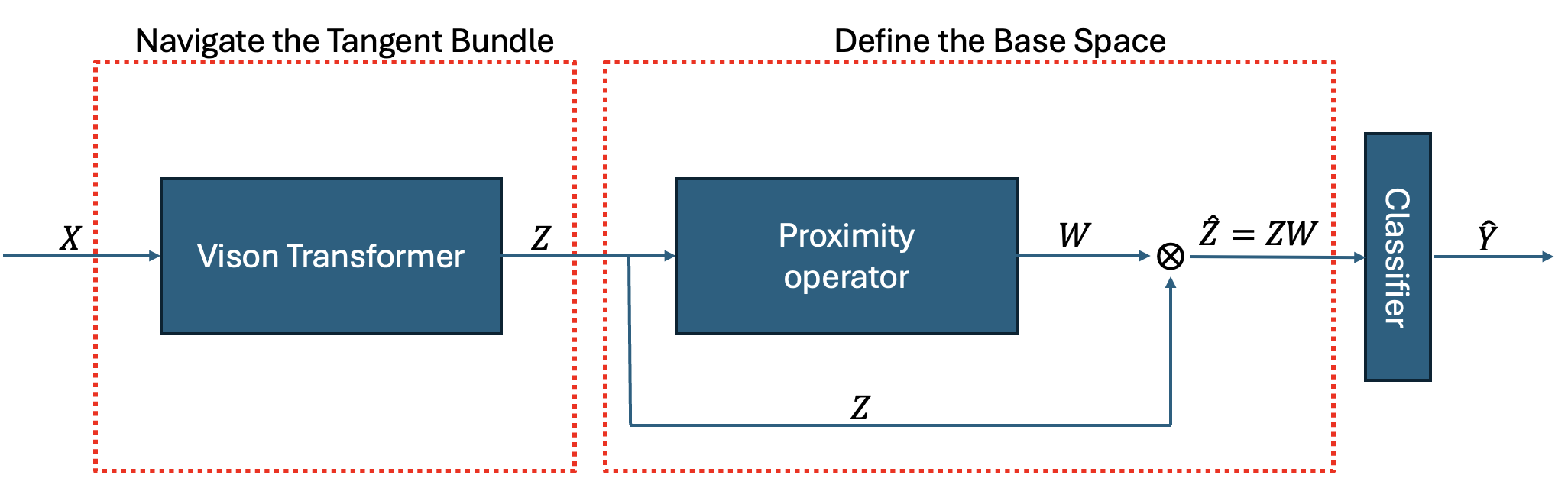}
  \caption{The overall architecture of the proposed model. The model consists of two core components: Vision Transformer (ViT) for feature extraction and proximity operator for geometric optimization.}
  \label{fig:model}
\end{figure}

This section introduces a novel model that improves upon the Vision Transformer (ViT) by incorporating proximal methods, which enhance feature representation and classification performance. The overall architecture of the model, as illustrated in Fig.~\ref{fig:model}, consists of two core components: the ViT and the proximity operator. These components operate in tandem within a unified geometric framework, forming the foundation of this innovative approach.

The ViT leverages an attention mechanism to extract features from input data and constructs a tangent bundle, where each attention head corresponds to a local tangent space. These tangent spaces capture the characteristics of the data from various local perspectives, yielding diverse high-dimensional feature representations. This tangent bundle structure serves as a critical basis for subsequent optimization steps.

The proximity operator is introduced as an improvement to the ViT, functioning to define a section within the tangent bundle and to project the data from the tangent spaces back onto the base space. Through this geometric optimization process, the proximity operator yields a refined self-representation, denoted as \( \hat{Z} = ZW \), where \( W \) represents the coefficient matrix obtained through optimization. This procedure ensures geometric consistency of the features while enhancing inter-class separability and intra-class compactness, thereby improving performance in classification tasks.

Subsequent subsections will elaborate on the relationship between the ViT and the tangent bundle, as well as the algorithmic rationale and geometric principles underlying the proximity operator, providing a detailed theoretical foundation for this novel model.

\subsection{Tangent Bundle Construction Through Vision Transformer}

Based on the concepts of manifolds and tangent spaces \cite{Lee00}, the Vision Transformer (ViT) employs a self-attention mechanism to construct a geometric representation of input data, thereby obtaining a tangent bundle. The input data \( X \) is assumed to lie on a manifold \( \mathcal{M} \),
a high-dimensional surface embedded in \( \mathbb{R}^D \),
where \( D \) denotes the ambient dimension of the flattened data
(not the number of patches in ViT).

The global structure of the manifold can be approximated through the tangent spaces \( T_x(\mathcal{M}) \) at specific points, such as local data points in the training set. In the ViT, the primary role of the self-attention mechanism is to construct and aggregate these tangent spaces to comprehensively model the structure of the manifold.

The process begins by dividing the input \( X \) into 
$n$ patches $\mathbf{x}_i \in \mathbb{R}^p$, where \( i \in \{1, 2, \dots, n\} \), which are projected into a high-dimensional space using a learnable embedding matrix \( \mathbf{E}\in \mathbb{R}^{d\times p} \)
generating vectors
\begin{equation}
\mathbf{z}_i = \mathbf{E} \mathbf{x}_i \in \mathbb{R}^d.
\label{eq:embedding}
\end{equation}
The embedded features \( (\mathbf{z}_i)_{1\le i \le n} \), together with a learnable class token \( \mathbf{z}_{\text{class}} \), are concatenated to form the initial input matrix \( \mathbf{Z}_0 \):
\begin{equation}
\mathbf{Z}_0 = [\mathbf{z}_{\text{class}}, \mathbf{z}_1, \mathbf{z}_2, \dots, \mathbf{z}_n]^\top \in \mathbb{R}^{(n+1) \times d}.
\label{eq:initial_input}
\end{equation}
In each self-attention layer $\ell$, token interactions are computed through the query, key, and value matrices
\begin{equation}
\mathbf{Q}_\ell = \mathbf{Z}_\ell \mathbf{W}_{\rm q}, \quad \mathbf{K}_\ell = \mathbf{Z}_\ell \mathbf{W}_{\rm k}, \quad \mathbf{V}_\ell = \mathbf{Z}_\ell \mathbf{W}_{\rm v},
\label{eq:qkv}
\end{equation}
where \( \mathbf{W}_{\rm q}, \mathbf{W}_{\rm k}, \mathbf{W}_{\rm v}\) are learnable weight matrices in $\mathbb{R}^{d \times d_{\rm k}}$, which are unique for each Transformer layer.

And \( \mathbf{Z}_\ell \) is the input to the \( \ell \)-th Transformer layer. The scaled dot-product attention is computed as
\begin{equation}
\mathbf{H}_\ell=\text{Attention}(\mathbf{Q}_\ell, \mathbf{K}_\ell, \mathbf{V}_\ell) = \underbrace{\text{softmax} \left( \frac{\mathbf{Q}_\ell \mathbf{K}_\ell^\top}{\sqrt{d_{\rm k}}} \right)}_{\displaystyle \mathbf{A}_\ell } \mathbf{V}_\ell.
\label{eq:attention}
\end{equation}
From a geometric perspective, the attention weights $\mathbf{A}_\ell$
define the relative importance of each token, effectively constructing the tangent space \( T_x(\mathcal{M}) \) for a given point \( x \), such as a patch or the class token. These tangent spaces capture the local variations of the manifold. By summing weighted contributions, the self-attention mechanism aggregates these tangent spaces into a unified representation.

The multi-head attention mechanism extends this process by computing tangent spaces in parallel across multiple heads:

\begin{equation}
\mathbf{H}_{\text{multi-head}} = 
[\mathbf{H}_1, \mathbf{H}_2, \dots, \mathbf{H}_h] \mathbf{W}_{\rm o},
\label{eq:multihead_attention}
\end{equation}
where, for every $j\in \{1,\ldots,h\}$, \( \mathbf{H}_j\ \) represents the output of the \( j \)-th attention head, \( h \) is the number of heads, and \( \mathbf{W}_{\rm o} \in \mathbb{R}^{hd_k \times d} \) is a learnable projection matrix. 

This parallel mechanism aggregates information from multiple tangent spaces, forming a local tangent bundle. When a sufficient number of attention heads is employed to capture the tangent spaces, the resulting tangent bundle can effectively approximate transformations across the entire manifold.

While the discussion above focuses on how the self-attention mechanism within each block builds the tangent bundle, it is important to note that the Vision Transformer includes additional operations following self-attention, such as residual connections, layer normalization, and a feedforward multilayer perceptron (MLP). These components further transform the intermediate output \( \mathbf{H}_{\text{multi-head}} \), enhancing the model’s expressiveness and enabling more effective feature refinement across layers. Specifically, the input to the next layer \( \mathbf{Z}_{\ell+1} \) 
is obtained through the standard Transformer block update following the multi-head attention:
\begin{align}
\mathbf{Z}_\ell' &= \mathbf{Z}_\ell + \mathbf{H}_{\text{multi-head}, \ell}, \label{eq:residual1} \\
\hat{\mathbf{Z}}_\ell &= \mathrm{LayerNorm}(\mathbf{Z}_\ell'), \label{eq:layernorm} \\
\mathrm{FFN}(\hat{\mathbf{Z}}_\ell) &= 
  \mathrm{GELU}(\hat{\mathbf{Z}}_\ell \mathbf{W}_{1,\ell} + \mathbf{b}_{1,\ell})\mathbf{W}_{2,\ell}+ \mathbf{b}_{2,\ell}, \label{eq:ffn} \\
\mathbf{Z}_{\ell+1} &= \mathbf{Z}_\ell' + \mathrm{FFN}(\hat{\mathbf{Z}}_\ell), \label{eq:residual2}
\end{align}
where the residual connections (Eqs.~\ref{eq:residual1} and \ref{eq:residual2}), 
layer normalization (Eq.~\ref{eq:layernorm}), 
and the feedforward multilayer perceptron (MLP, Eq.~\ref{eq:ffn})  
progressively refine the intermediate representation, 
enhancing both expressiveness and feature consistency across layers.

As multiple layers of the self-attention mechanism are stacked, the ViT incrementally refines the tangent bundle representation. In the final layer, the class token \( \mathbf{z}_{\text{class}}^{(L)} \) aggregates global information from the manifold and serves as a descriptor of the overall structure of the input data. In typical classification tasks, only the final class token \( \mathbf{z}_{\text{class}}^{(L)} \) is used as the global representation, while the patch embeddings \( \mathbf{z}_1^{(L)}, \dots, \mathbf{z}_n^{(L)} \) are discarded.
This enables the ViT to comprehensively represent the geometric properties of the input, achieving robust performance in downstream tasks such as classification.

It is worth emphasizing that the proposed method is not equivalent to directly applying sparsification or optimization on the projection matrix \( \mathbf{W}_{\rm o} \) in Equation~\eqref{eq:multihead_attention}. Specifically, our approach operates exclusively on the class token, rather than processing all patch tokens as \( \mathbf{W}_{\rm o} \) does, which contributes to reducing both the parameter count and computational complexity. In addition, the patch embeddings transformed by \( \mathbf{W}_{\rm o} \) are subsequently passed through multiple nonlinear operations, leading to a fundamental difference in the type of information being processed. Therefore, the proposed method cannot be interpreted as a simple sparsification of \( \mathbf{W}_{\rm o} \). Further methodological details are provided in the next section.

\subsection{Geometric Optimization via Learnable Proximity Operator}

We propose a proximal method that leverages geometric structures to optimize the representations obtained from the Vision Transformer (ViT). The model is grounded in concepts such as manifolds, tangent bundles, and sections. In this framework, the ViT constructs a tangent bundle, where each attention head corresponds to a tangent space encoding local feature representations. The proximity operator is used to define a section on this tangent bundle and project features onto the base space, thereby ensuring geometrically consistent self-representations.

In the basic formulation, we begin with the following optimization objective:
\begin{equation}
\label{eq:objective_function}
\min_{\mathbf{W} \in C} \|\mathbf{Z} - \hat{\mathbf{Z}}_{\mathbf{W}}\|_{\rm F}^2 + g(\mathbf{W}),
\end{equation}
where \( \mathbf{Z} \in \mathbb{R}^{d \times m} \) denotes the batch-wise aggregation of final-layer class tokens \( \mathbf{z}_{\text{class}}^{(L)} \) extracted from the ViT, where each column vector in \( \mathbf{Z} \) corresponds to a single class token \( \mathbf{z}_{\text{class}}^{(L)} \) from one input sample in the batch, \( \hat{\mathbf{Z}}_{\mathbf{W}} = \mathbf{Z}\mathbf{W} \) represents their self-representation, and \( \mathbf{W} \) is the coefficient matrix to be optimized. Geometrically, \( \mathbf{Z} \) can be viewed as a collection of points on the tangent bundle, while \( \hat{\mathbf{Z}}_{\mathbf{W}} \) represents their projections onto the base space through the section defined by the proximity operator. The term \( g(\mathbf{W}) \) imposes structural regularization, such as sparsity \cite{7178678}.

The gradient of the reconstruction loss \( f(\mathbf{W}) = \frac{1}{2} \|\mathbf{Z} - \mathbf{Z}\mathbf{W}\|_{\rm F}^2 \) is given by
\begin{equation}
\label{eq:gradient_error}
\nabla f(\mathbf{W}) = \mathbf{Z}^\top \mathbf{Z} (\mathbf{W} - \mathbf{I}_m),
\end{equation}
Based on this, we apply the proximity operator to perform the optimization according to a proximal-gradient scheme:
\begin{equation}
\label{eq:proximal_update}
\mathbf{W}_{k+1} = \operatorname{prox}_{\gamma_kg + \iota_C} \left( \mathbf{W}_k - \gamma_k\mathbf{Z}^\top \mathbf{Z} (\mathbf{W}_k - \mathbf{I}_m) \right),
\end{equation}
This update effectively refines \( \mathbf{W} \) toward the optimal solution while preserving geometric alignment across the tangent bundle.

Building upon this formulation, we further enhance the optimization process by introducing a learnable matrix \( \mathbf{R}_k \) that adaptively modulates the descent direction and step size. The updated iteration becomes:
\begin{equation}
\label{eq:learnable_proximal_update}
\mathbf{W}_{k+1} = \operatorname{prox}_{\gamma_kg + \iota_C} \left( \mathbf{W}_k - \gamma_k\mathbf{Z}^\top (\mathbf{Z}\mathbf{W}_k - \mathbf{Z}) \mathbf{R}_k \right),
\end{equation}
where \( \mathbf{R}_k \in \mathbb{R}^{m \times m} \) is a learnable transformation matrix. This matrix serves as a data-driven unmatched preconditioner and is inspired by quasi-Newton methods\cite{berahas2021quasinewtonmethodsmachinelearning}, which approximate second-order curvature information to accelerate convergence. In our model, \( \mathbf{R}_k \) functions as a learnable quasi-Newton approximation, enabling fast and effective updates.

\begin{figure}[ht]
  \centering
  \includegraphics[scale = 0.35]{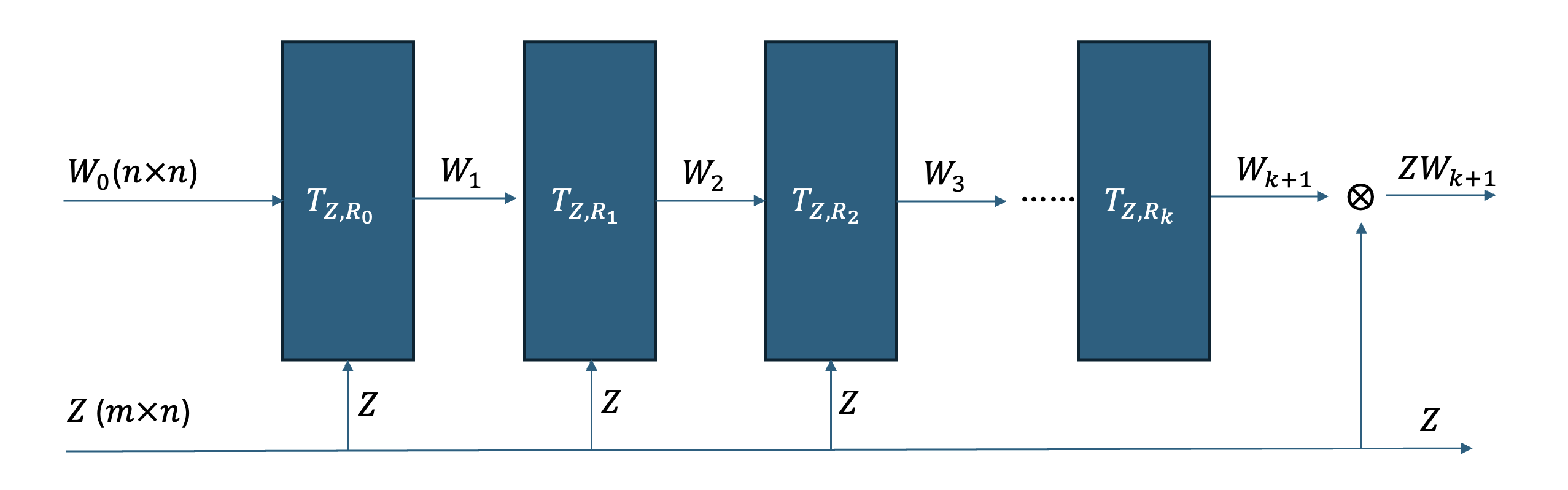}
  \caption{Illustration of the iterative updates in the proposed proximal method. $T_{Z,R_i}$ denotes the proximity operator applied to the input feature matrix $Z$ 
with the parameter matrix $R_i$ at the $i$-th iteration. 
The sequence $W_0, W_1, \dots, W_{k+1}$ represents the iteratively updated weights,
and the final output is $Z W_{k+1}$.
}
  \label{fig:prox_structure}
\end{figure}

Geometrically, the learnable proximity operator dynamically adjusts how the class token representations are projected onto the base space, improving intra-class compactness and inter-class separability. As in the original design, we apply soft-thresholding followed by a ReLU activation to ensure sparsity and non-negativity in \( \mathbf{W} \). The structure of this learnable proximal module is shown in Fig.~\ref{fig:prox_structure}.

The entire optimization procedure, which includes tangent space extraction, bundle construction, section definition, and learnable projection, is summarized in Algorithm~\ref{alg:proposed_model}. This optimization process aligns the class tokens and generates an optimized self-representation.

\begin{algorithm}[H]
\caption{Proposed Model}
\label{alg:proposed_model}
\KwIn{Input data $X$, learnable parameters of ViT $\theta_{\text{ViT}}$ and classifier $\theta_{\text{clf}}$, 
initial matrix $\mathbf{W}_0$, iteration number $k_{\max}$
}

\KwOut{Predicted labels $\hat{Y}$}

Divide $X$ into patches $x_i$ and project each patch into a high-dimensional space using embedding matrix $\mathbf{E}$: \\
$\mathbf{z}_i \gets \mathbf{E} x_i$\;

Concatenate patch embeddings with the class token: \\
$\mathbf{Z}_0 \gets [\mathbf{z}_{\text{class}}, \mathbf{z}_1, \mathbf{z}_2, \dots, \mathbf{z}_n]$\;

\For{each Transformer layer $\ell$}{
    $\mathbf{Q} \gets \mathbf{Z}_\ell \mathbf{W}_{\rm q}$, $\mathbf{K} \gets \mathbf{Z}_\ell \mathbf{W}_{\rm k}$, $\mathbf{V} \gets \mathbf{Z}_\ell \mathbf{W}_{\rm v}$\;
    
    $\mathbf{A} \gets \text{softmax}\left(\frac{\mathbf{Q} \mathbf{K}^\top}{\sqrt{d_k}}\right)$, \quad $\mathbf{H} \gets \mathbf{A} \mathbf{V}$\;
    
    $\mathbf{Z}_{\ell+1} \gets \text{NonlinearUpdate}(\mathbf{Z}_\ell, \mathbf{H})$\;
}

Collect final-layer class tokens from all samples in the batch: \\
$\mathbf{Z} \gets [\mathbf{z}_{\text{class}}^{(L,1)}, \mathbf{z}_{\text{class}}^{(L,2)}, \dots, \mathbf{z}_{\text{class}}^{(L,m)}]$;

Initialize $\mathbf{W}_0$ and set $k \gets 0$\;

\While{$k < k_{\max}$}{
    $\nabla f(\mathbf{W}_k) \gets \mathbf{Z}^\top (\mathbf{Z} \mathbf{W}_k - \mathbf{Z})$\;
    
    $\mathbf{U}_k =\mathbf{W}_k - \gamma_k \nabla f(\mathbf{W}_k) \mathbf{R}_k$\;
    
    $\mathbf{W}_{k+1} = \max\left(0, \mathrm{sgn}(\mathbf{U}_k) \cdot \max(|\mathbf{U}_k| - \gamma_k\lambda, 0)\right)$\;
    
    $k \gets k + 1$\;
}

$\hat{\mathbf{Z}}_{\mathbf{W}} \gets \mathbf{Z} \mathbf{W}_{k+1}$\;

$\hat{Y} \gets \text{Classifier}(\hat{\mathbf{Z}}_{\mathbf{W}}; \theta_{\text{clf}})$\;

\Return $\hat{Y}$
\end{algorithm}

\section{Experiments}
\label{experiments}

\subsection{Results on Various Datasets}

We conducted extensive experiments on four benchmark datasets to evaluate the effectiveness of the proposed model: Flowers, 15-Scene~\cite{Oliva2001ModelingTS}, Mini-ImageNet~\cite{5206848}, and CIFAR-10~\cite{Krizhevsky2009Learning}.

These datasets vary in both scale and resolution: Flowers and 15-Scene are relatively small but high-resolution datasets, Mini-ImageNet is a large-scale high-resolution dataset (we used the $224 \times 224$ high-resolution version), and CIFAR-10 is a low-resolution dataset.

To ensure fair comparisons and minimize the impact of training procedures and hardware configurations, all experiments were initialized using the same pre-trained Vision Transformer (ViT) backbone and trained under consistent settings. Specifically, the ViT model follows the standard configuration, consisting of 12 Transformer blocks with 12 attention heads per block and an embedding dimension of 768. 

We evaluated three configurations: (1) the original ViT; (2) ViT integrated with a non-learnable proximal method; and (3) ViT enhanced with a learnable proximal method, which incorporates trainable step sizes $\gamma_k$ and preconditioning matrices $R_k$.

As shown in Table~\ref{tab:results}, both variants of the proximal method consistently improve classification accuracy across all datasets. Notably, the learnable version significantly improves accuracy while also reducing the number of required iterations—achieving comparable or better performance with approximately half the number of unfolded layers compared to its fixed counterpart. This demonstrates that incorporating learnable parameters not only enhances model performance but also accelerates convergence during training.

Moreover, we observed that performance gains are more pronounced on high-resolution datasets (e.g., Flowers, 15-Scene, and Mini-ImageNet), while moderate but consistent improvements are still evident on low-resolution datasets like CIFAR-10.

These results collectively validate the robustness and generalizability of the proposed method under diverse data conditions, demonstrating its effectiveness across various visual recognition tasks.

\begin{table}[ht]
\centering
\renewcommand{\arraystretch}{1.4}
\setlength{\tabcolsep}{4pt}
\resizebox{\textwidth}{!}{%
\begin{tabular}{c|c|c|ccccc}
\hline
\textbf{Dataset} & \textbf{Class} & \textbf{Batch size} & \textbf{ACC (ViT)} & \textbf{ACC (ViT+Prox)} & \textbf{Imp. to ViT} & \textbf{ACC (ViT+LearnableProx)} & \textbf{Imp. to ViT} \\ \hline

Flowers          & 5              & 64                  & 98.1\%             & 99.9\%                 & \textbf{1.8\%}                & 99.9\%                           & \textbf{1.8\%}                \\ 
\hline
15-Scene         & 15             & 64                  & 97.4\%             & 99.6\%                 & 2.2\%                & 99.8\%                           & \textbf{2.4\%}                \\
\hline
Mini-ImageNet    & 100            & 128                 & 95.7\%             & 97.8\%                 & 2.1\%                & 98.1\%                           & \textbf{2.4\%}                \\ 
\hline
CIFAR-10         & 10             & 64                  & 97.8\%             & 98.2\%                 & 0.4\%                & 98.5\%                           & \textbf{0.7\%}                \\ 
\hline
\end{tabular}%
}
\caption{Performance Comparison of ViT, ViT+Prox, and ViT+LearnableProx on Various Datasets}
\label{tab:results}
\end{table}

\subsection{Impact of Data Distribution}
Compared to using ViT alone, the proposed model significantly improves the distribution of feature representations for classification tasks, particularly when a learnable proximal method is incorporated. Intra-class distances are reduced, resulting in lower variance within each class, while inter-class distances are increased, leading to clearer separation between classes. These improvements are attributed to the enhanced regularization and structural learning capabilities provided by the learnable proximal components.

To evaluate these effects, two types of analyses are conducted: t-SNE visualizations and Wasserstein distance comparisons. These analyses collectively highlight improvements in both intra-class compactness and inter-class separability.

\subsubsection{Visualization of the data distribution}

To demonstrate the impact of the proposed model on the structure of feature representations, t-SNE~\cite{JMLR:v9:vandermaaten08a} is employed to visualize the distributions in a 2D space. t-SNE is a dimensionality reduction algorithm that maps high-dimensional data to a low-dimensional space while preserving local neighborhood relationships. Specifically, given $N$ high-dimensional data points $\mathbf{x}_1, \mathbf{x}_2, \ldots, \mathbf{x}_N$, the pairwise similarity in the original space is defined using the conditional probability $p_{ij}$:

\begin{equation}
p_{j|i} = \frac{\exp(-\|\mathbf{x}_i - \mathbf{x}_j\|^2 / 2\sigma_i^2)}{\sum_{k \neq i} \exp(-\|\mathbf{x}_i - \mathbf{x}_k\|^2 / 2\sigma_i^2)}, \quad p_{ij} = \frac{p_{j|i} + p_{i|j}}{2N}.
\label{eq:high_dim_similarity}
\end{equation}
In the low-dimensional space, the similarity between the embedded points $\mathbf{y}_1, \mathbf{y}_2, \ldots, \mathbf{y}_N$ is defined as

\begin{equation}
q_{ij} = \frac{(1 + \|\mathbf{y}_i - \mathbf{y}_j\|^2)^{-1}}{\sum_k \sum_{l \neq k} (1 + \|\mathbf{y}_k - \mathbf{y}_l\|^2)^{-1}}.
\label{eq:low_dim_similarity}
\end{equation}
The goal of t-SNE is to minimize the Kullback-Leibler (KL) divergence~\cite{shlens2014noteskullbackleiblerdivergencelikelihood} between the high-dimensional similarity distribution $P$ and the low-dimensional distribution $Q$:

\begin{equation}
\text{KL}(P \parallel Q) = \sum_{i \neq j} p_{ij} \log \frac{p_{ij}}{q_{ij}},
\label{eq:kl_divergence}
\end{equation}
so that the low-dimensional layout faithfully preserves the local structure of the high-dimensional data.

\begin{figure}[ht]
  \centering
  \includegraphics[scale=0.23]{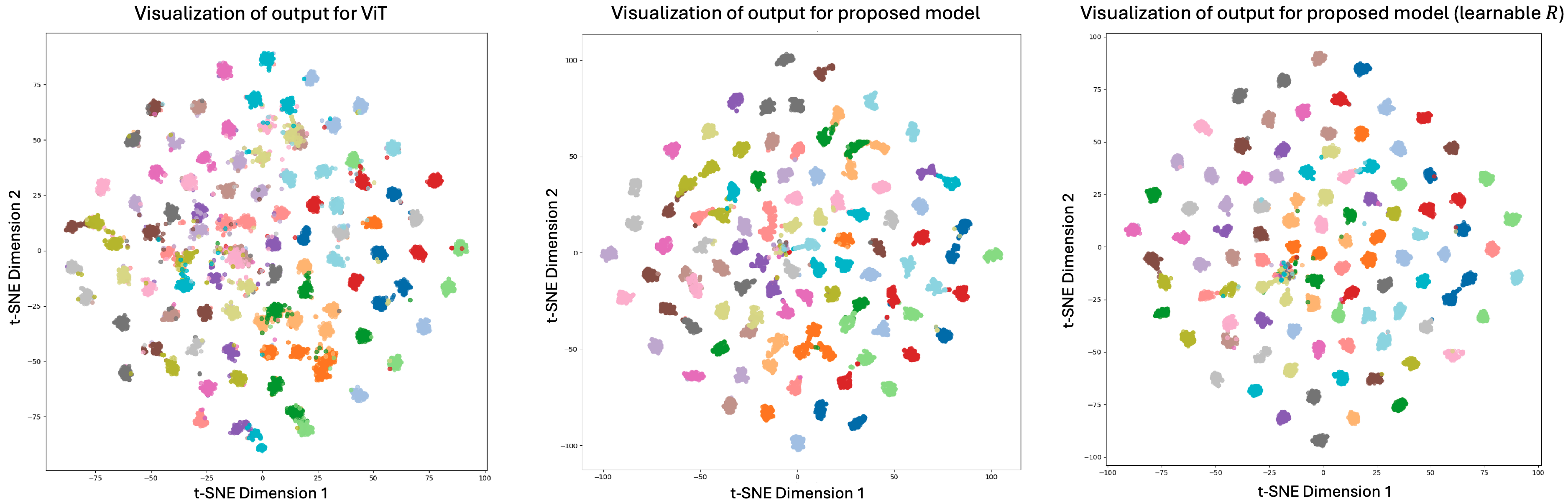}
  \caption{t-SNE visualization of feature distributions on the Mini-ImageNet dataset. 
From left to right: (a) ViT only; (b) ViT + non-learnable proximal method; 
(c) ViT + learnable proximal method. 
Colors are randomly assigned to different classes within each subfigure to distinguish the classes, and the same color across subfigures does not correspond to the same class.
}
  \label{fig:tsne}
\end{figure}

Figure~\ref{fig:tsne} illustrates the two-dimensional t-SNE embeddings of the extracted features on the Mini-ImageNet dataset under three different configurations. In subfigure (a), which corresponds to ViT alone, the data points within each class are loosely distributed, with significant variance and numerous outliers scattered across regions dominated by other classes. This dispersed structure indicates weak intra-class compactness and overlapping between classes, which may lead to ambiguous decision boundaries during classification. The result after applying the non-learnable proximal method is shown in subfigure (b). Compared to the baseline, most class clusters become more compact, and the number of outliers is significantly reduced. The overall feature distribution appears more structured, and the overlap between different classes is alleviated, demonstrating the effectiveness of this optimization strategy in improving feature organization. Subfigure (c) illustrates the outcome achieved using the learnable proximal method. The improvement is even better: samples within each class are tightly grouped, the number of outliers is significantly reduced, and the separation between different classes becomes much clearer. This more structured distribution reflects enhanced intra-class consistency and improved inter-class separability, confirming that the learnable formulation enables the model to better capture discriminative patterns in the feature space.

\subsubsection{Wasserstein Distance for Inter-Class Separation}

To measure inter-class distances, the Wasserstein distance (also known as the Earth Mover’s Distance)\cite{Panaretos_2019} is employed. The Wasserstein distance quantifies the discrepancy between two distributions by calculating the minimum transportation cost required to transform one distribution into the other. For discrete data, the Wasserstein distance is defined as:

\begin{equation}
W_1(\mu, \nu) = \min_{\gamma \geq 0} \sum_{i=1}^{P} \sum_{j=1}^{Q} \gamma_{ij} C_{ij}, 
\label{eq:wasserstein_distance}
\end{equation}
where \( \mu = \{x_i\}_{i=1}^{P} \) and \( \nu = \{y_j\}_{j=1}^{Q} \) represent two discrete probability distributions, 
\( C_{ij} \) denotes the cost or distance between \( x_i \) and \( y_j \), which is chosen as the Euclidean distance \( C_{ij} = \| x_i - y_j \|_2 \), 
and \( \gamma_{ij} \) is the transport flow between \( x_i \) and \( y_j \). 
The transport matrix \( \gamma \) must satisfy the following constraints:

\begin{equation}
\sum_{j=1}^{Q} \gamma_{ij} = \mu_i,\quad \forall i\in\{1,\dots,P\},
\label{eq:row_constraint}
\end{equation}
\begin{equation}
\sum_{i=1}^{P} \gamma_{ij} = \nu_j,\quad \forall j\in\{1,\dots,Q\},
\label{eq:column_constraint}
\end{equation}

where the row sums of \( \gamma \) equal the weights of distribution \( \mu \), and the column sums equal the weights of distribution \( \nu \).

By leveraging the Wasserstein distance, the inter-class discrepancies can be effectively quantified, providing a reliable metric for further analysis.

The class-wise Wasserstein distance matrix \( A \) is used to represent the distances between each class. \( A \) is a symmetric matrix, where \( a_{ij} \in A \) denotes the distance between class \( i \) and class \( j \). Each element \( a_{ij} \) is calculated using Equation \eqref{eq:wasserstein_distance}.

\begin{figure}[ht]
  \centering
  \includegraphics[scale = 0.42]{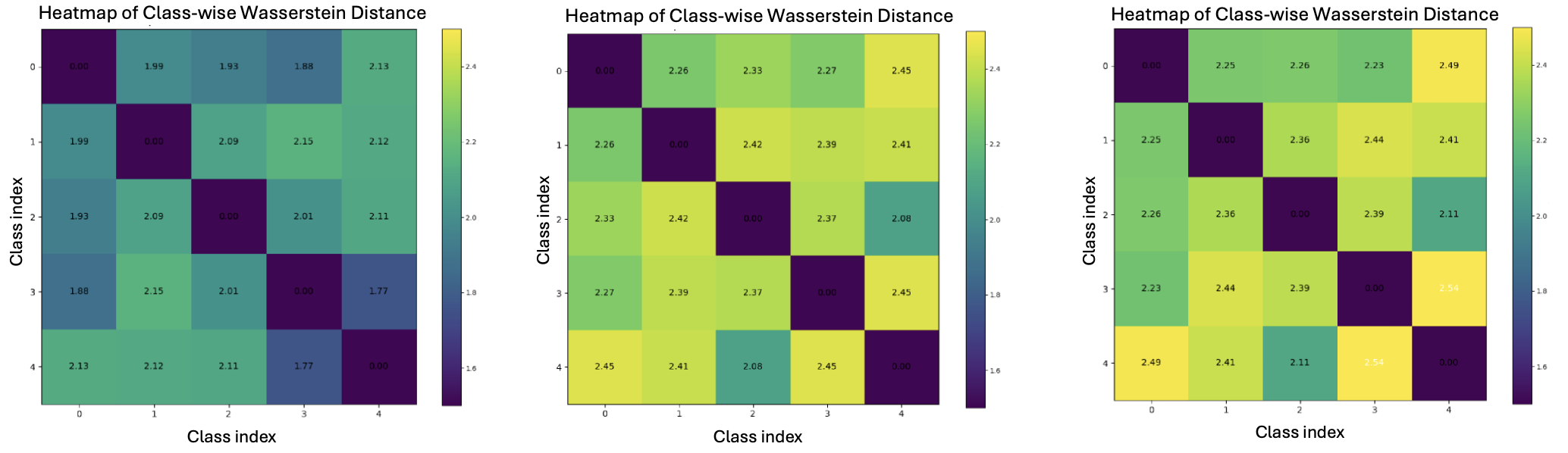}
  \caption{Class-wise Wasserstein distance matrix on the Flowers dataset. 
(a) Matrix generated by the baseline ViT, showing smaller inter-class distances and higher similarity between class distributions. 
(b) Matrix generated by the proposed model with the non-learnable proximal method, showing increased inter-class distances and more distinct distributions. 
(c) Matrix generated by the proposed model with the learnable proximal method, achieving the most pronounced inter-class separation.
}
  \label{fig:w-f}
\end{figure}

\begin{figure}[ht]
  \centering
  \includegraphics[scale = 0.26]{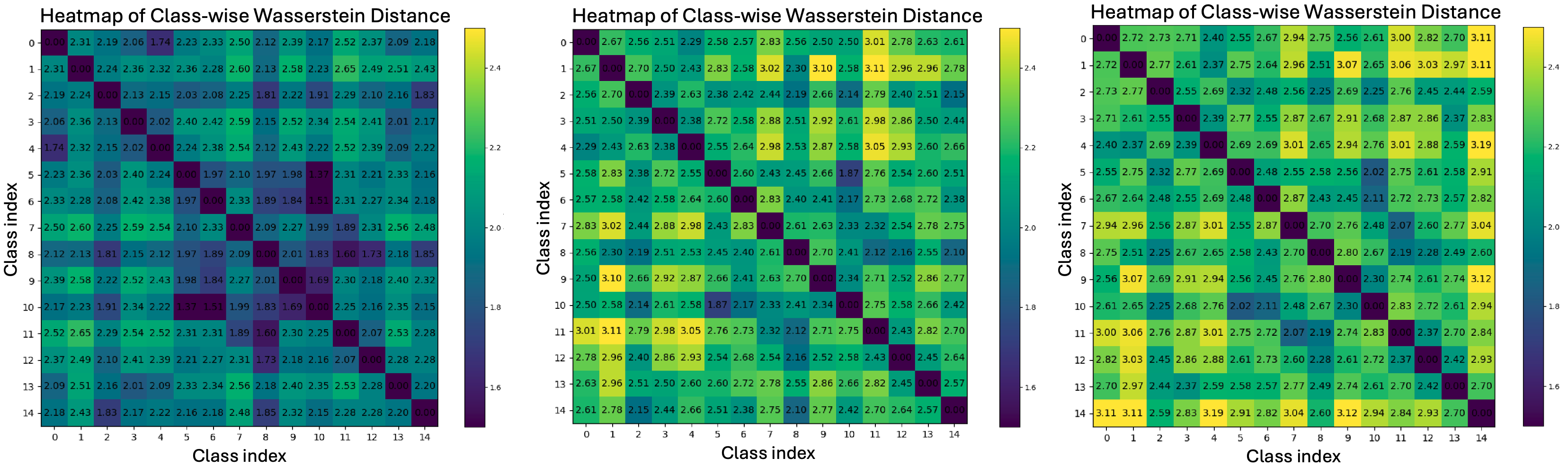}
  \caption{Class-wise Wasserstein distance matrices on the 15-Scenes dataset.  
(a) Matrix from the baseline ViT, with less distinct separation and higher similarity between class distributions.  
(b) Matrix from the proposed model with the non-learnable proximal method, exhibiting improved inter-class separability.  
(c) Matrix from the proposed model with the learnable proximal method, showing the clearest and most structured separation among classes.
}
  \label{fig:w-s}
\end{figure}

Figures~\ref{fig:w-f} and~\ref{fig:w-s} show the Wasserstein distance matrices for two datasets: Figure~\ref{fig:w-f} corresponds to the Flowers dataset, and Figure~\ref{fig:w-s} corresponds to the 15-Scene dataset. Each figure includes results from three model configurations: the original ViT, ViT with a non-learnable proximal method, and ViT with a learnable proximal method. In each matrix, the brightness of a cell reflects the Wasserstein distance between the corresponding pair of classes—brighter cells indicate greater dissimilarity and stronger inter-class separability, while darker cells suggest that the distributions are more similar or even overlapping.

A consistent pattern can be observed across both datasets. The matrices produced by the original ViT exhibit low overall brightness and weak contrast between classes, indicating limited feature discriminability. After introducing the non-learnable proximal gradient descent algorithm, the matrices become noticeably brighter, and the distances between many class pairs are significantly increased. With the learnable proximal method, the matrices achieve the highest brightness and clearest boundaries between classes, accompanied by a stronger overall contrast. These results demonstrate the progressive improvements in modeling inter-class distances and confirm the effectiveness of the proposed method in enhancing classification performance and feature discrimination.

\subsubsection{Proximity Operator Placement Frequency}

In our default setting, the unrolled proximal algorithm is applied only after the final ViT block to refine the class token representation.
However, it is also possible to insert the proximal operator at intermediate Transformer blocks, which may influence the learned feature distributions.

To further investigate the impact of where and how often the proximal processing is applied within the Vision Transformer (ViT), two complementary experiments are conducted on the Mini-ImageNet dataset: one varies the specific block at which the operator is applied, and the other varies the number and frequency of insertions.

As shown in Table~\ref{tab:proximal_summary}, applying the proximal processing at different depths leads to varying classification performance. When inserted in the early layers (e.g., 2nd block), the performance is limited (95.1\%). This suggests that at such shallow stages, the learned features are not yet semantically meaningful, and enforcing a projection may disrupt natural representation learning. Intermediate blocks (e.g., 6th or 8th) yield moderate improvements, whereas placing the operator in later blocks, especially the final one (12th), achieves the highest accuracy (98.1\%). This is attributed to the fact that deeper layers capture more stable and well-structured tangent bundles, making the class token better aligned with the underlying semantic submanifold.

In addition to the placement, we examine the effect of applying the proximity operator at multiple stages simultaneously. Table~\ref{tab:proximal_summary} (right) shows that while moderate gains can be observed when applying it to multiple blocks (e.g., blocks 1, 6, 12 or 1–12), the most effective setup remains applying it only after the final block. This observation reinforces the hypothesis that excessive or premature enforcement of structure may inject undesirable directional bias, leading to suboptimal final representations despite correction attempts in later layers.

In summary, applying the proximity operator solely at the final block of the Vision Transformer yields the best performance. This is attributed to the stability and structure of deep-layer features, which enable the projection to more effectively align with the semantic submanifold, while avoiding the directional bias introduced by early-stage interventions.

\begin{table}[t]
\centering
\begin{subtable}[t]{0.45\textwidth}
\centering
\vspace{0pt}  
\begin{tabular}{@{}lc@{}}
\toprule
\textbf{Block Position} & \textbf{Accuracy (\%)} \\
\midrule
None  & 95.7 \\
After Block 2     & 95.1 \\
After Block 4     & 95.5 \\
After Block 6     & 95.5 \\
After Block 8     & 96.1 \\
After Block 10    & 96.8 \\
After Block 12    & \textbf{98.1} \\
\bottomrule
\end{tabular}
\caption*{(a) Single Insertion of Proximity Operator}
\end{subtable}
\hfill
\begin{subtable}[t]{0.45\textwidth}
\centering
\vspace{0pt}  
\begin{tabular}{@{}lc@{}}
\toprule
\textbf{Blocks Applied} & \textbf{Accuracy (\%)} \\
\midrule
None                         & 95.7 \\
Blocks 1, 6, 12              & 97.6 \\
Blocks 1, 2, 6, 7, 12        & 97.6 \\
Blocks 1, 2, 6, 7, 9, 10, 12 & 97.7 \\
All Blocks (1–12)            & 97.9 \\
Only Block 12                & \textbf{98.1} \\
\bottomrule
\end{tabular}
\caption*{(b) Multiple Insertions of Proximity Operator}
\end{subtable}

\vspace{2mm}
\caption{Classification accuracy (\%) on Mini-ImageNet when applying proximity operator to different Transformer blocks. (a) shows performance with a single proximity operator inserted at various stages. (b) shows accuracy when proximity operator is applied to multiple blocks.}
\label{tab:proximal_summary}
\end{table}

\section{Conclusion}
\label{conclusion}

This paper presents a novel framework that integrates the Vision Transformer (ViT) with a proximal optimization strategy to address ViT’s limitations in modeling global geometric structures. By combining the self-attention mechanism of ViT, which constructs a tangent bundle over the data manifold, with proximal operations that define sections and project data onto the base space, the proposed method achieves joint geometric regularization.

Extensive experiments on several representative datasets validate the effectiveness and broad applicability of the proposed framework. Compared to the standard Vision Transformer (ViT), the approach achieves significant improvements in classification accuracy and feature distribution quality without introducing substantial model complexity. In particular, it demonstrates stronger intra-class compactness and inter-class separability, especially on high-resolution image tasks. The results indicate that integrating geometric optimization principles into deep neural architectures can enhance global structural modeling while preserving the benefits of end-to-end training. This work not only advances the theoretical depth of ViT in feature optimization but also offers a new perspective for building more robust and interpretable visual models.

Overall, this work underscores the effectiveness of combining geometric insights with Transformer-based architectures for learning discriminative and robust feature representations.

\section{Acknowledgments}

This work was supported by the Army Research Office (ARO) under Grant No.~W911NF2410329. This work was prepared as an account of research sponsored by the U.S.~Government. Neither the U.S.~Government nor any agency thereof assumes any legal liability or responsibility for the accuracy or usefulness of the information contained herein. The views and opinions expressed are those of the authors and do not necessarily reflect those of the U.S.~Government or any agency thereof.

\newpage
\bibliographystyle{unsrt}
\bibliography{reference.bib}
\end{document}